\documentclass[letterpaper, 10 pt, conference]{ieeeconf}
\pdfoutput=1
\usepackage{graphicx}
\usepackage{amsmath}
\usepackage{amssymb}
\usepackage{floatrow}
\usepackage{multirow}
\usepackage{adjustbox}
\newfloatcommand{capbtabbox}{table}[][\FBwidth]
\usepackage[dvipsnames]{xcolor}
\usepackage{blindtext}
\usepackage{cite}
\usepackage{siunitx}
\usepackage[format=plain,labelfont={bf,it},textfont=it,font=small,belowskip=-1.5pt]{caption} 

\usepackage[colorlinks]{hyperref}
\newcommand{\sota}{state-of-the-art}
\newcommand{\algoname}{\textit{ProxEmo}}
\title{\LARGE \textbf{\algoname}:  Gait-based Emotion Learning and \\Multi-view Proxemic Fusion for Socially-Aware Robot Navigation}

\author{Venkatraman Narayanan, Bala Murali Manoghar, Vishnu Sashank Dorbala, Dinesh Manocha, and Aniket Bera \\
{University of Maryland, College Park, USA}\\
{\small{Supplemental version including Code, Video, Datasets at \url{https://gamma.umd.edu/proxemo/}}}\vspace{-15pt}
}

\begin{document}

\maketitle
\thispagestyle{empty}
\pagestyle{empty}

\begin{abstract}

We present \textit{\algoname}, a novel end-to-end emotion prediction algorithm for socially aware robot navigation among pedestrians. Our approach predicts the perceived emotions of a pedestrian from walking gaits, which is then used for emotion-guided navigation taking into account social and proxemic constraints. To classify emotions, we propose a multi-view skeleton graph convolution-based model that works on a commodity camera mounted onto a moving robot.
Our emotion recognition is integrated into a mapless navigation scheme and makes no assumptions about the environment of pedestrian motion.
It achieves a mean average emotion prediction precision of $82.47\%$ on the Emotion-Gait benchmark dataset. We outperform current state-of-art algorithms for emotion recognition from 3D gaits.
We highlight its benefits in terms of navigation in indoor scenes using a Clearpath Jackal robot.

\end{abstract}

~\vspace{-0.4cm}
\section{Introduction}
Recent advances in AI and robotics technology are gradually enabling humans and robots to coexist and share spaces in different environments. This is especially common in places such as hospitals, airports, and shopping malls.
Navigating a robot with collision-free and socially-acceptable paths in such scenarios poses several challenges \cite{social5}. For example, in the case of a crowded shopping mall, the robot needs to be aware of the intentions of an oblivious shopper coming towards it for friendly navigation. Knowing the perceived emotional state of a human in such scenarios allows the robot to make more informed decisions and navigate in a socially-aware manner.

Understanding human emotion has been a well-studied subject in several areas of literature, including psychology, human-robot interaction, etc. There have been several works that try to determine the emotion of a person from verbal (speech, text, and tone of voice) \cite{speechemo,textemo} and non-verbal (facial expressions, walking styles, postures) \cite{face2emo,pose2emo} cues. There also exist multi-modal approaches that use a combination of these cues to determine the person's emotion \cite{multiemo2,multiemo1,multiemo3}.

In our work, we focus on emotionally-aware robot navigation in crowded scenarios. Here, verbal cues for emotion classification are not easily attainable. With non-verbal cues, facial expressions that are often occluded from the egocentric view of the robot and might not be fully visible. Besides, emotion analysis from facial features is a topic of debate in several previous works: these features are inherently unreliable caused by vague expressions emerging from a variety of psychological and environmental factors~\cite{face_emo_drawback,face_emo_drawback2}. As such, in our work, we focus on using ``walking styles" or ``gaits" to extract the emotions of people in crowds.

\begin{figure}[t!]
    \centering
    \includegraphics[width=\linewidth]{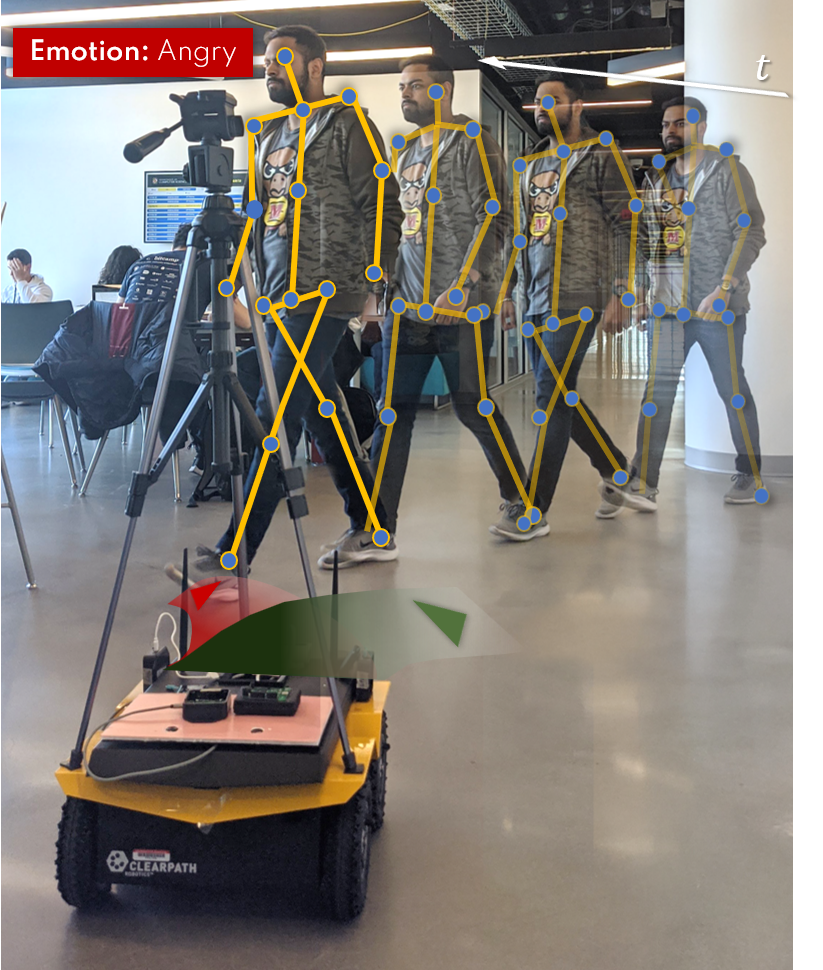}
        \caption{\textbf{ProxEmo}:  We present a gait-based emotion and proxemics learning algorithm to perform socially-aware robot navigation. The \textcolor{red}{\textbf{red}} arrow indicates the path of the robot without social awareness. The \textcolor{OliveGreen}{\textbf{green}} arrow indicates the new path after an \textit{angry} emotion is detected. Observe the significant shift away from the pedestrian when an \textit{angry} gait is detected. This form of navigation is especially useful when the robot is expected to navigate safely through crowds without causing discomfort to nearby pedestrians.
        \vspace{-0.5cm}}
    \label{fig:Emotion}
    \vspace{-0.1cm}
\end{figure}

\begin{figure*}[t]
\centering
\includegraphics[width=1\textwidth]{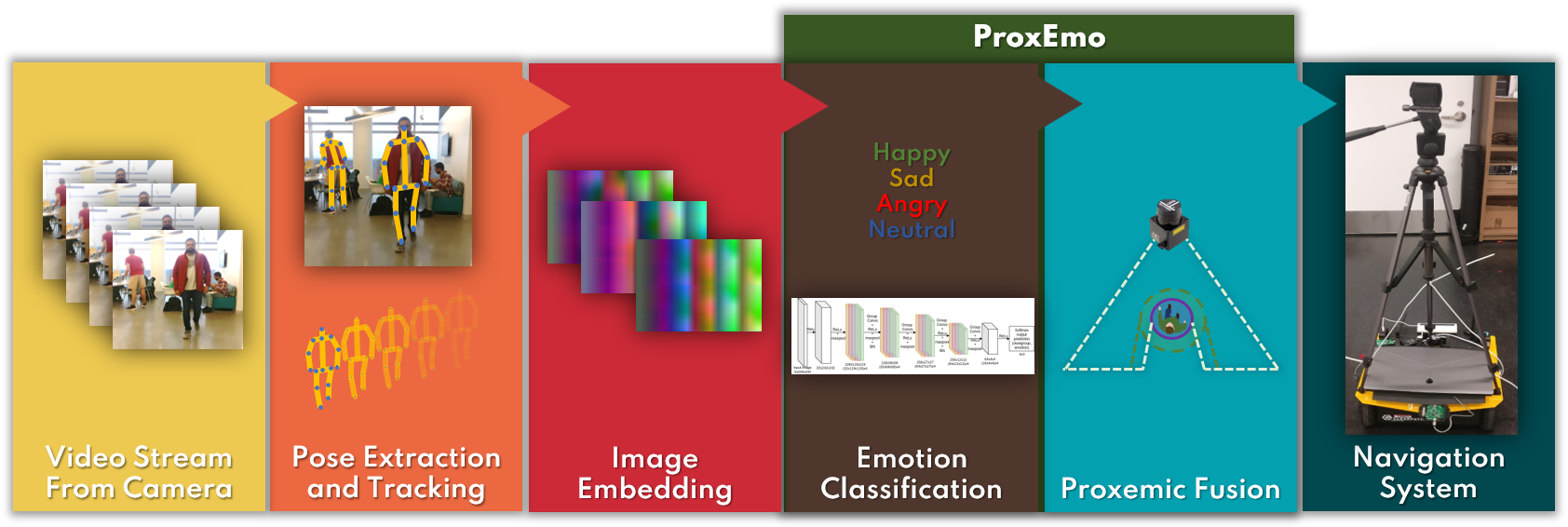}
\caption{\textbf{Overview of our Pipeline}: We first capture an RGB video from an onboard camera and extract pedestrian poses and track them at each frame. These tracked poses over a predefined time period are embedded into an image, which is then passed into our {\algoname} model for classifying emotions into four classes. The obtained emotions then undergo \textit{proxemic fusion} with the LIDAR data and are finally passed into the navigation stack.
\vspace{-0.5cm}}
\label{fig:overview}
\end{figure*}

Obtaining perceived emotions from gaits is a challenging problem that has been well documented in the past. More recently, various machine learning solutions \cite{mobilegait},\cite{bhattacharya2020step} have been proposed to tackle this problem. However, these approaches suffer from the following drawbacks: 
\vspace{-3pt}
\begin{itemize}
    \item The training datasets used are singular in direction, i.e., there is motion capture only when a person is walking in a straight line towards the camera. This is a significant disadvantage for our task of socially-aware crowd navigation, where the robot often encounters people walking from several directions towards or away from the camera.
    \item Some approaches that are tailored towards using emotion for enhancing the task of robot navigation assume a static overhead camera that captures the trajectories of pedestrians. This is not ideal, as the overhead camera might not always be available in all scenarios.
\end{itemize}
To overcome these challenges, we propose \textit{\algoname}, a novel algorithm for realtime gait-based emotion classification for socially-guided navigation. {\algoname} is tailored towards working with commodity RGB egocentric cameras that can be retrofitted onto moving platforms or robots for navigating among pedestrians. The \textbf{major contributions} of our work can be summarized as follows:
\begin{itemize}
    \item We introduce a novel approach using group convolutions to classify pedestrian emotions from gaits, which drastically improves accuracy compared to SOTA.
    \item Our method explicitly takes into consideration pedestrian behavior in crowds as we train our model on skeletal data of people approaching the robot from multiple directions, as opposed to approaching from a single view from the front.
    \item We present a new navigation scheme using \textbf{\emph{Proxemic Fusion}} that accounts for pedestrian emotions. 
    \item Finally, we introduce a \textbf{\textit{Variational Comfort Space}}, which integrates into our navigation scheme, taking into account varying pedestrian orientations.
\end{itemize}

We note that identifying the true nature of a person's emotion via only a visual medium can be difficult. Therefore in this work, we focus only on the \emph{perceived} emotions from the point of an external observer as opposed to \emph{actual} internal emotion. 

\section{Related Work}
\label{Sec:Related_Work}
In this section, we present a brief overview of social-robot navigation algorithms. We also review related work on emotion modeling and classification from visual cues.

\subsection{Social Robotics and Emotionally-Guided Navigation}
As robots have become more commonplace, their impact on humans' social lives has emerged as an active area of research.
Studies from multiple domains \cite{social1,social2, bera2016glmp, chandra2020cmetric} have tried to quantify this impact in several ways.
In \cite{social5}, Kruse et al. present a comprehensive survey on navigation schemes for robots in social scenarios. They describe various social norms (interpersonal distances, human comfort, sociability) that the robot must consider not to cause discomfort to people around it.
Michaid et al. \cite{social6} discuss about how robots can attain \emph{artificial} emotions for social interactions. Several classical \cite{wilkie2009generalized,van2010optimal,van2008reciprocal} and deep learning \cite{loquercio2018dronet} approaches tackle the problem of navigation through highly dynamic environments. More recently, reinforcement learning methods \cite{long2018towards,fan2018crowdmove} have been described for collision avoidance in such environments. For pedestrian handling, in particular, Randhavane et al. \cite{randhavane2019pedestrian} make use of a pedestrian dominance model (PDM) to identify the dominance level of humans and plan a trajectory accordingly. 
In \cite{socialsurvey}, Rios-Martinez et al. present a detailed survey on the proxemics involved with socially aware navigation. In \cite{infospace}, Kitazawa et al. discuss ideas such as \textit{Information Process Space} of a human.
In \cite{socaware}, Pandey et al. discuss a strategy to plan a socially aware path using milestones.

\subsection{Emotion Modeling and Classification}
There exists a substantial amount of research that focuses on identifying the emotions of humans based on body posture, movement, and other non-verbal cues.
Ruiz-Garcia et al. \cite{face2emotion1} and Tarnowski et al. \cite{face2emotion2}, use deep learning to classify different categories of emotion from facial expressions.  The approach by \cite{multiemo3} uses multiple modalities such as facial cues, human pose and scene understanding. Randhavane et al.~\cite{randhavane2019identifying, randhavane2019liar} classify emotions into four classes based on affective features obtained from 3D skeletal poses extracted from human gait cycles. Their algorithm, however, requires a large number of 3D skeletal key-points to detect emotions and is limited to single individual cases. Bera et al. \cite{bera2019emotionally,bera2020fg} classify emotions based on facial expressions along with a pedestrian trajectory obtained from overhead cameras. Although this technique achieves good accuracy in predicting emotions from trajectories and facial expressions, it explicitly requires overhead cameras in its pipeline.

\subsection{Action Recognition for Emotions}
\label{action_comparison}
The task of action recognition involves identifying human actions from sequences of data (usually videos) \cite{ActionRecogSurvey}.
A common task in many of these models is recognizing gait-based actions such as walking and running.
Thus, the task of gait action recognition is closely related to the task of emotion recognition from gaits, as both perform classification on the same input. Bhattacharya et al. \cite{bhattacharya2020step,bhattacharya2019take} use graph convolutions for the emotion recognition task, in a method similar to the action recognition model used in Yan et al. \cite{yan2018spatial}. Ji et al. \cite{ji2019large} propose a CNN-based method that gives state-of-the-art results on gait based action recognition tasks. Their model is invariant to viewpoint changes.

\section{Overview and Methodology}
\label{Sec:Methodology}
We propose a novel approach, {\algoname}, for classifying emotions from gaits that works with an egocentric camera setup. Our method uses 3D poses of human gaits obtained from an onboard robot camera to classify perceived emotions. These perceived emotions are then used to compute variable proxemic constraints in order to perform socially aware navigation through a pedestrian environment.\\
Figure \ref{fig:overview} illustrates how we incorporate {\algoname} into an end-to-end \textit{emotionally-guided} navigation pipeline. 

 The following subsections will describe our approach in detail. We first discuss the dataset and the augmentation details we used for training. Then, we briefly discuss our pose estimation model, followed by a detailed discussion of our emotion classification model, ~\algoname. Finally, we describe how socially-aware navigation can be performed using the obtained emotions.
 
\subsection{Notations} \label{SubSec:notation}
In our formulation, we represent the human with $16$ joints as shown in figure \ref{fig:skeleton_temporal}. Thus, a pose $P \in \mathbb{R}^{16\times3}$ of a human is a set of 3D positions of each joint $j_i$, where $i \in \{0,1, ..., 15\}$. For any RGB video $V$, we represent the gait extracted using 3D pose estimation as $G$. The gait $G$ is a set of 3D poses ${P_1, P_2,..., P_{\tau}}$ where $\tau$ is the number of frames in the input video $V$. 

\subsection{Dataset Preparation} \label{SubSec:data_prep}
\begin{figure}[h]
\centering
\includegraphics[width=1.0\textwidth]{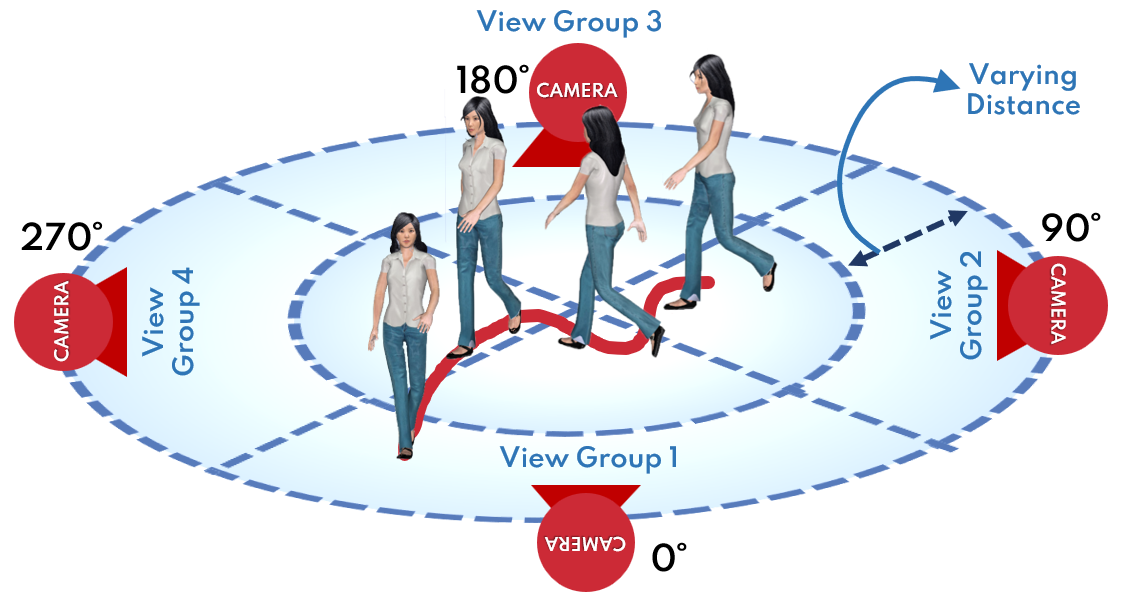}
\caption{\textbf{Data Augmentation}: \textit{By applying specific translations and rotations, we augment the data into different camera views. We divide the viewpoints into four view-groups based on the angle of approach to categorize the direction in which the person is walking. The augmentations also take into consideration varying distances of the camera from the origin point of the gait sequence. }}
\label{fig:data_aug}
\end{figure}

We make use of two labeled datasets by Randhavane et al.~\cite{Ewalk} and Bhattacharya et al.~\cite{bhattacharya2020step}, containing time-series 3D joints of $342$  and $1835$ gait cycles each (a total of $2177$ gait samples). Each gait cycle has $75$ timesteps with $16$ joints as shown in Figure \ref{fig:skeleton_temporal}. Thus, each sample in this new dataset has a dimension of $joints \times time \times$ dimensions $= 16*75*3 = 3600$. These samples are labeled into $4$ emotion classes: \textit{angry}, \textit{sad}, \textit{happy}, and \textit{neutral} with $10$ labelers per video (to capture the perceptual difference between different labelers).
In order to train our network for prediction from multiple views, we augment the dataset as follows. First, we consider a camera placed at a particular distance from the human, as shown in Figure \ref{fig:data_aug}. Then for different camera positions oriented towards the human, we perform augmentation by applying transformations given in equation \ref{eq1}.

\begin{align}
\label{eq1}
j_{aug}
=
\begin{bmatrix}
cos \theta & 0 & -sin \theta \\
0 & 1 & 0 \\
sin \theta & 0 & cos \theta
\end{bmatrix}
\times
\begin{bmatrix}
x\\
y\\
z
\end{bmatrix}
+
\begin{bmatrix}
T_x\\
T_y\\
T_z
\end{bmatrix}
\end{align}

where $j_{aug}$  are the coordinates of the augmented joints, $T_x, T_y , T_z$ are the translation vectors, and $\theta$ is the rotation along $Y$ axis. For our experiments, we attain $72\times4 = 288$ augmentations for each sample by considering $\theta$ at gradients of $5^{\circ}$, with 4 translations of $[1m-4m]$ along the $Z$ axis ($T_{z}$). Thus, after augmentation, we have a total of $288\times2177 = 626,976$ gaits in our dataset.

\subsection{Human-Pose Estimation}\label{sec:poseEstimation}
\begin{figure}[h]
\centering
\includegraphics[width=1.0\textwidth]{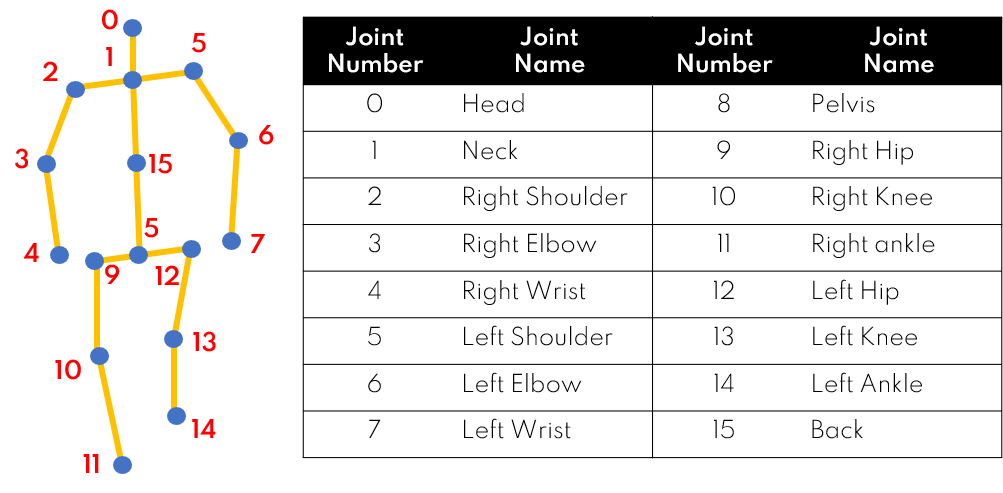}
\caption{\textbf{Skeleton Representation}: We represent a pedestrian by $16$ joints ($j_i$). The overall pose of the pedestrian is defined using these joint positions.
\vspace{-0.4cm}}
\label{fig:skeleton_temporal}
\end{figure}

A pose estimation strategy for humans walking in a crowded real-world scenario has to be robust to noise coming from human attire or any items they might be carrying. To account for this, we employ a robust approach described in \cite{dabral2018learning} for this task. Their paper describes a two-step network trained in a weakly supervised fashion. First, a \textit{Structure-Aware PoseNet (SAP-Net)} trained on spatial information provides an initial estimate of joint locations of people in video frames. Later, a \textit{Temporal PoseNet (TP-Net)} trained on time-series information corrects this initial estimate by adjusting illegal joint angles and joint distances. 
The final output is a sequence of well-aligned 3D skeletal poses P. Figure \ref{fig:skeleton_temporal} is a representation of the skeletal output obtained.
\subsection{Generating Image Embeddings}
\label{sec:image}
\begin{figure*}[h]
\centering
\includegraphics[width=1.0\textwidth]{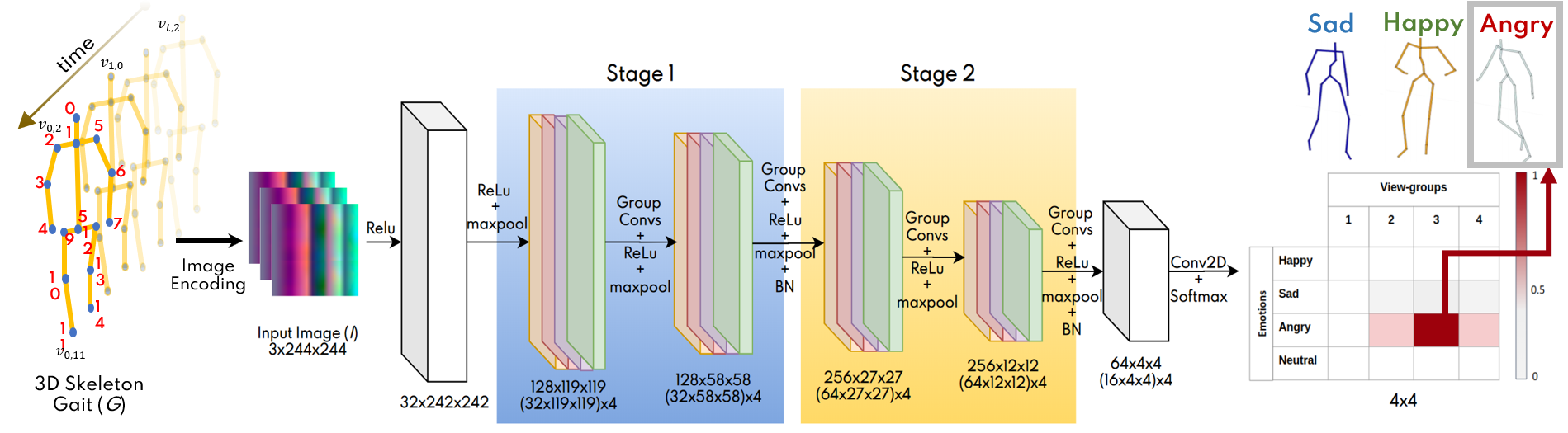}
\caption{\textbf{{\algoname} Network Architecture}: The network is trained on image embeddings of the 5D gait set G, which are scaled up to $244\times244$. The architecture consists of four group convolution (GC) layers. Each GC layer consists of four groups that have been stacked together. This represents the four group convolution outcomes for each of the four emotion labels. The group convolutions are stacked in two stages represented by \textcolor{blue}{\textbf{Stage 1}} and \textcolor{Dandelion}{\textbf{Stage 2}}. The output of the network has a dimension of $4\times4$ after passing through a $softmax$ layer. The final predicted emotion is given by the maxima of this 4$\times$4 output.
\vspace{-0.5cm}}
\label{fig:vs_gcnn}
\end{figure*}

We observe that 2D convolutions are much faster and efficient as opposed to graph convolutions \cite{ji2019large}.
Hence, we embed the spatial-temporal skeletal gait sequence $G$ as an image $I$, using the equations described in \ref{eqn:image}.

\begin{align}
\label{eqn:image}
I &= \{R_{(x,y)} = Z_{(t,j)} ; G_{(x,y)} = Y_{(t,j)}; B_{(x,y)} &= X_{(t,j)}\}
\end{align}

Here, $R$, $B$, and $G$ are image color channels, $x, y$ are the co-ordinates of image, and $X_{(t,j)}, Y_{(t,j)}, Z_{(t,j)}$ are the co-ordinates of skeletal joint $j$ at time $t$.This image $I$ is finally upscaled to $244\times244\times3$ for training our {\algoname} model. 

\subsection{\textbf{\algoname}: Classifying Emotions from Gaits}
Figure~\ref{fig:vs_gcnn} illustrates the architecture of our model for emotion classification. 
The image embedding $I$ obtained from the gaits are passed through two stages of group convolutions to obtain an emotion label.

\subsubsection{Group Convolutions}
We take inspiration from ~\cite{krizhevsky2012imagenet} and make use of group convolutional layers in designing our {\algoname} architecture. \textbf{Group Convolution Layers} ($GC$), in essence, operate just like 2D convolution layers, except that they fragment the input into $n_g$ groups and perform convolution operations individually on them before stacking the outputs together.
The advantage of doing this is that the network learns from different parts of the input in isolation.
This is especially useful in our case because we have a dataset that varies based on two factors, view-group and emotion labels. 
The variation in the view-groups is learned by the different convolution groups $GC$, and the emotions are learned by the convolutions taking place within each group.
Group convolutions increase the number of channels in each layer by $n_g$ times. The output ($h_{i}$) of each group in the convolution layer is $h_{i} = x_{i} * k_{i}$ and $h_{out} = [h_{1} |...| h_{n_g}]$.
 where, $h_{out}$ is the output of the group convolution, $x_{i}$ represents the input, and $k_{i}$ represents the kernel for convolution. The output $[h_{1} |...| h_{n_g}]$ is a matrix concatenation of all the group outputs along channel axis. In our case, we choose $n_g$ as 4 because we have $4$ view-groups.
\subsubsection{{\algoname} Architecture}
The network consists of seven convolution layers. The initial layer is a traditional 2D convolution layer, which performs channel up-sampling for the forthcoming group convolution operations. These operations take place in two stages -\\ \textbf{Stage 1}: This consists of two \textit{GC} layers, each having 128 convolution filters (32 per group $\times~ n_g$).\\
\textbf{Stage 2}: This consists of two convolution \textit{GC} layers, however, unlike stage 1, each \textit{GC} 256 convolution filters (64 per view-group $\times~n_g$).\\
Both traditional 2D convolution and \textit{GC} layers are passed through a $ReLU$ non-linear activation and max pooling layer. \\
The outputs from \textbf{Stage 1} and \textbf{Stage 2} are represented by $h_{s}$ where $s={1, 2}$. We also perform \emph{batch normalization}. The output of each both the group convolution stages, $h_s$ are given by,
\begin{align}
    p_{s}^{*} &= GC(x_{s}, k_{s}^{1}) \notag \\
    p_{s} &= MaxPool(ReLU(p_{s}^{*})) \notag \\
    h_{s}^{*} &= GC(p_{s}, k_{s}^{2}) \notag \\
    h_{s} &= BatchNorm(MaxPool(ReLU(h_{s}^{*}))) 
\end{align}

where, $s$ represents the two group convolution stages as described before, $x_{s}$ is the input to the group convolution stage `$s$', $k_{s}^{1}$ and $ k_{s}^{2}$ represent convolution kernels for first and second $GC$ layers within a stage, $p_{s}^{*}$ and $h_{s}^{*}$ are the first and second $GC$ layer outputs determined using equation above.
 
After performing the group convolutions, the output $h_{2}$ is passed through two 2D convolution layers. These convolution layers help in gathering the features learned by the $GC$ layers to finally predict both the view-group and emotion of the gait sequences. \\ 

Rather than using fully-connected layers for predicting the view-group, our method utilizes convolution layers to predict the $n_k \times n_g$ output, where $n_k$ is the number of emotions and $n_g$ is the number of view-groups. This makes our model considerably lighter (number of model parameters) and faster (run-time performance), compared to other {\sota} algorithms. 


The final output of the classifier consists of multi-class $softmax$ prediction, $E_{i, j}$, given by the equation \ref{softmax_calc}. Here $e_{i, j}$ refers to the final hidden layer output of the network, where $i=0, 1,\dots (n_k -1)$ is the emotion class and $j=0, 1, \dots, (n_g-1)$ is the view-group class.

\begin{equation}\label{softmax_calc}
    E_{i, j} = \frac{\exp{(e_{i, j})}}{\sum_{i=0}^{n_k-1}\sum_{j=0}^{n_g-1}\exp{(e_{i, j})}}
\end{equation}

$E_{i,j}$ can be considered as a $4\times 4$ matrix containing $16$ values corresponding to different view-groups and emotions.

\subsection{Emotion-guided Navigation using Proxemic Fusion}
\label{Sec:Navigation}
\begin{figure}[h]
\centering
\includegraphics[width=0.9\textwidth]{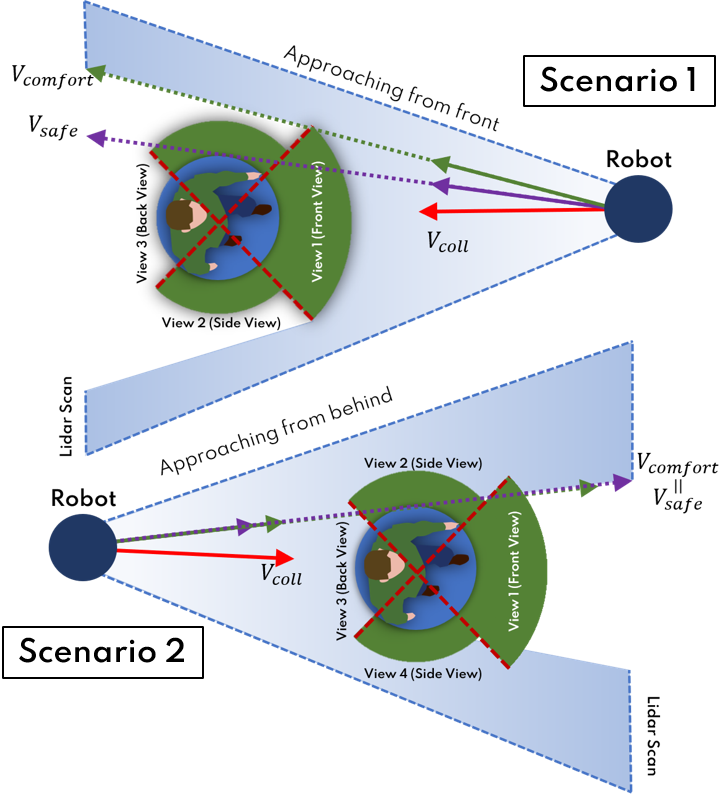}
    \caption{
    \textbf{Variational Comfort Space}:
    We consider a varying comfort space $c$ around a person based on their position (defined by the view-group $g$) in front of the robot. In scenario 1, the pedestrian approaches the robot from the front. Here, as the pedestrian is aware of the robot's presence, it needs to be more respectful of the proxemic comfort space and take action $V_{comfort}$ represented by the \textcolor{OliveGreen}{green} arrow.
    In scenario 2, the robot is approaching the person from behind. An unaware pedestrian need not be disturbed by the robot, due to which it can be more liberal with its actions. The \textcolor{violet}{violet} arrow representing the safe action $V_{safe}$ coincides with $V_{comfort}$ in this case.
    \vspace{-0.5cm}
    }
\label{fig:human_lidar}
\end{figure}

We use the emotions $E_{i,j}$ predicted from {\algoname} to compute the comfort space ($c$) of a pedestrian, which is the socially comfortable distance (in cm) around a person.
We combine $c$ along with the LIDAR data ($L$) to perform ``\emph{proxemic fusion}'' (\ref{proxemic_fusion}), obtaining a set of points where it is permissible for the robot to navigate in a \emph{socially-acceptable} manner.
This is illustrated in figure \ref{fig:human_lidar}.\\
\vspace{-0.2cm}
\subsubsection{Comfort Space Computation}
In order to model the predicted emotions $E_{i,j}$ from {\algoname} into a comfort space distance $c$, we use the following equation:
\begin{align}\label{eq_socdist}
c = \frac{\sum_{j=1}^{4} c_j \cdot \max{(E_j)} }{\sum_{j=1}^{4}{E_j}} \cdot v_g
\end{align}
Here, $E_{j}$ represents a column vector of the \textit{softmax} output, which corresponds to the group outcomes for each individual emotion. $c_j$ is a constant derived from psychological experiments described in~\cite{ruggiero2017effect} to compute the limits on an individual's comfort spaces and is chosen from a set $\{90.04, 112.71, 99.75, 92.03\}$ corresponding to the comfort spaces (radius in cm) for \{\textit{happy, sad, angry, neutral}\} respectively. We acknowledge that these distances depend on many factors, including cultural differences, environment, or a pedestrian's personality, and restrict our claims to variations in comfort spaces due to the emotional difference. These distances are based on how comfortable pedestrians are while interacting with others. $v_g$ is a view-group constant defined in the following subsection.

 \subsubsection{Variational Comfort Space ($v_g$)}
 \label{sec:vg_space}
 We take inspiration from the \emph{Information Process Space} defined by Kitazawa et al. \cite{infospace} to define our own Variational Comfort Space constant $v_g$. This constant acts as a scaling factor in the comfort space based on the orientation of the pedestrian in the robot's view. This orientation is easily obtainable as {\algoname} also gives us a view-group output along with the emotion.\\
$v_g$ is chosen from a set of $\{1, 0.5, 0, 0.5\}$ based on the view group $g$ predicted. 
This is chosen based on the fact that people have varying personal space with respect to their walking direction, i.e., a pedestrian will care more about his/her personal space in front as compared to the sides. Also, the pedestrian might not care about personal and comfort space behind them since it does not lie in their field of view~\cite{kim2014personal}.\\
In figure \ref{fig:human_lidar}, we look at two scenarios to illustrate how the robot handles pedestrians considering variational comfort spaces:
\begin{itemize}
    \item \textbf{Scenario 1}: The robot is positioned in front of the person walking towards it. This is classified as view-group 1, having a $v_g$ value of 1. As the robot is visible to the person, in this case, it should be more precautious in safely maneuvering around the person. The comfort space around the pedestrian is larger in this case, and the robot takes a more skewed trajectory.
    \item \textbf{Scenario 2}: The robot is approaching the pedestrian from behind. This gait is classified as view-group 3 and has a $v_g$ value of 0. As the robot is not in the person's field of vision, in this case, it can afford to safely pass around the fixed space $F_s$ of the person.
\end{itemize}

At any time instant, the velocity of the robot will be directed towards the goal, and if there is an obstacle, it will lead to a collision $v_{coll}$. If an obstacle avoidance algorithm is used, the navigation scheme avoids it with an action $v_{safe}$. However, for socially acceptable proximally-aware navigation, this is not sufficient, as this requires the robot to follow certain social norms.
In order to adhere to these social norms, we incorporate the emotions predicted by {\algoname} to navigate in a socially acceptable manner represented by $V_{comfort}$.

\subsubsection{Proxemic Fusion}
\label{proxemic_fusion}
We fuse the LIDAR data to include proxemic constraints by performing a Minkowski sum ($M$) of the set of LIDAR points $L$ and a set containing the points in a circle $Z$ defined by a radius $r$.
The Minkowski sum $M$ provides us with a set of all the admissible points where the robot can perform emotionally-guided navigation. This is formulated using the following equations.
\begin{align}\label{eq_msum}
L &= \{a \: | \: a-a_{0} = d_{lidar}\}\\
Z &= \{b \: | \: dist(a,b) \leq r\} \notag \\
M \: &= \: L + Z = \{ a + b \: | \: a \in L, b \in Z \} \notag
\end{align}

Here, $a_{0}$ is a reference point on the LIDAR, and $d_{lidar}$ is the distance measurement (in metres). $r$ is the \textit{inflation radius} and is defined using the comfort space $c$ as:
\begin{align}
    r \: = \: c - [max(dh) - min(dh)]
\end{align}
where $dh \in L$ is a set of the LIDAR distances only for points where a human was detected. 
The maximum value of $dh$ corresponds to the farthest distance from the person from their fixed inner space $F_s$, while the minimum value of $dh$ corresponds to the closest distance of the person from this space. $F_s$ is represented by the blue circle around the person in the figure \ref{fig:human_lidar}. In terms of mathematical morphology, the outcome of \textit{proxemic fusion} is similar the dilation operation of the human, modelled as a obstacle, with the comfort space.

\begin{figure}[h]
\centering
\includegraphics[width=1\textwidth]{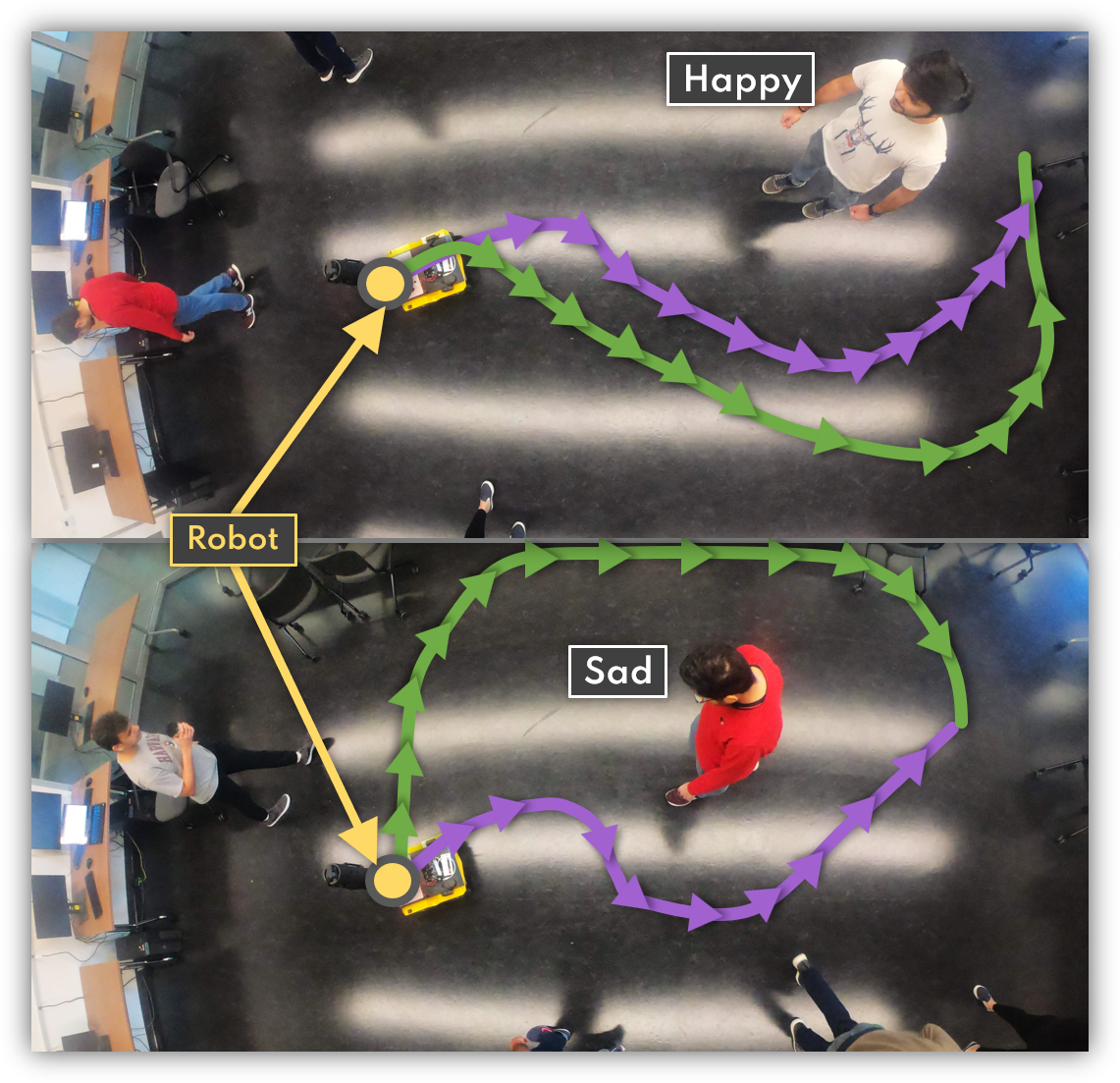}
    \caption{
    \textbf{Emotionally-Guided Navigation}: We use the emotions detected by {\algoname} along with the LIDAR data to perform \textit{Proxemic Fusion}. This gives us a comfort distance $c$ around a pedestrian for \textit{emotionally-guided} navigation. The \textcolor{OliveGreen}{green} arrows represent the path after accounting for $c$ while the \textcolor{violet}{violet} arrows indicate the path without considering this distance. Observe the significant change in the path taken in the \emph{sad} case. Note that the overhead image is representational, and {\algoname} works entirely from a egocentric camera on a robot.
    \vspace{-0.4cm}}
\label{fig:human_navigationres}
\end{figure}

\begin{figure*}[t!]
    \centering
    \includegraphics[width=\linewidth]{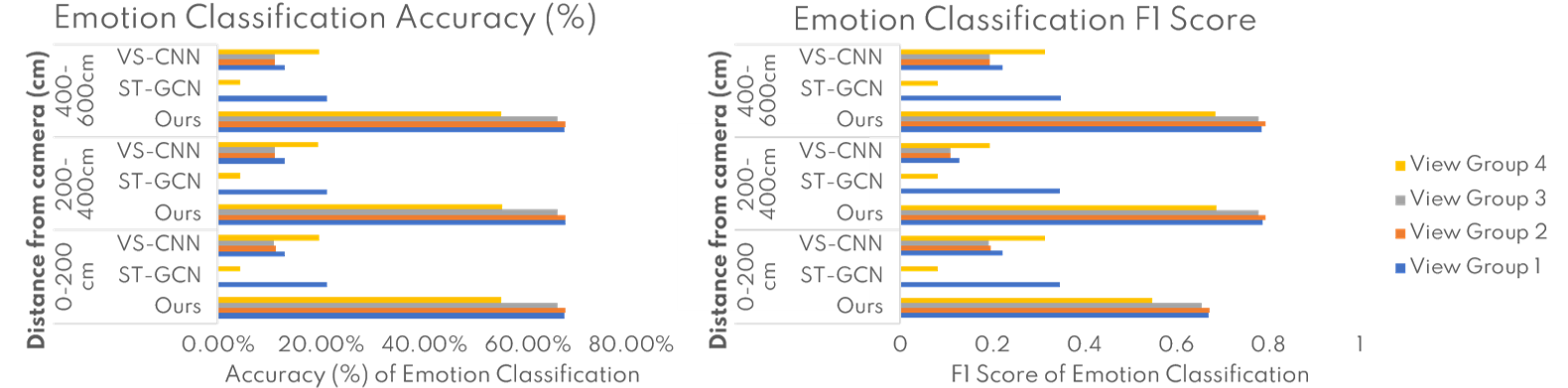}
        \caption{\textbf{Comparison of \algoname~with other arbitrary view algorithms }:  Here we present the performance metrics (discussed in section \ref{sec:eval_metrics}) of our {\algoname}~network compared to the \sota~arbitrary view action recognition models. We perform a comprehensive comparison of models across multiple distances of skeletal gaits from the camera and across multiple view-groups. It can be seen that our {\algoname}~network outperforms other \sota~network by \textbf{\textit{50\%}} at an average in terms of prediction accuracy.
        \vspace{-0.5cm}}
    \label{fig:model_acc_results}
    \vspace{-0.1cm}
\end{figure*}

\section{Experiments and Results}
\label{Sec:Results}

\subsection{Evaluation Metrics}\label{sec:eval_metrics}
We evaluate our model using two metrics:
\begin{itemize}
    \item \textbf{Mean Accuracy (\%)} - $\frac{1}{n_k \times n_g}\sum_{i=0}^{n_k}\sum_{j=0}^{n_g} \frac{TP_{i, j}}{N_{i,j}}$ \\
    \item \textbf{Mean F1 score} - $\frac{2}{n_k \times n_g}\sum_{i=0}^{n_k}\sum_{j=0}^{n_g} \frac{Pr_{i, j} * Rc_{i, j}}{Pr_{i, j} + Rc_{i, j}}$
\end{itemize}
where, $n_k~(=4)$ is the number of emotion classes, $n_g~(=4)$ is the number of view-groups, $TP_{i, j}$ is the number of true predictions for $i^{th}$ emotion class and $j^{th}$ view-group, $N_{i, j}$ is the total number of data samples for $i^{th}$ emotion class and $j^{th}$ view-group,
$Pr_{i, j}$ and $Rc_{i, j}$ is the \textit{precision} and \textit{recall} for $i^{th}$ emotion class and $j^{th}$ view-group. All the metrics mentioned are derived from a \textit{confusion matrix} generated by comparing actual vs predicted emotion and view-group for the data samples.

\begin{figure}[h]
    \centering
    \includegraphics[width=0.9\columnwidth]{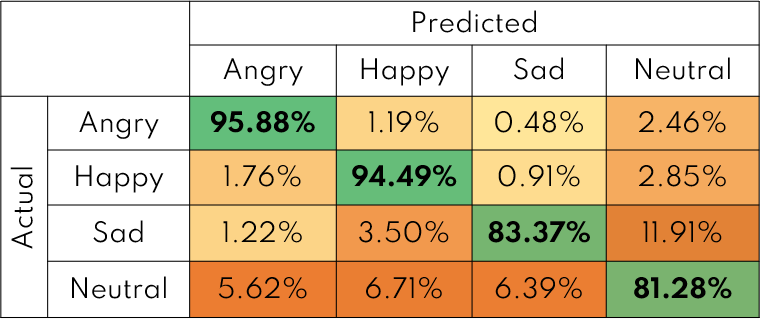}
    \caption{\textbf{Confusion Matrix}: We show the percentage of gaits belonging to every emotion class that were correctly classified by our algorithm,~\algoname.
    \vspace{-0.7cm}}
    \label{fig:cm}
\end{figure}

\subsection{Implementation Details} \label{SubSec:eval_dataset}

For training, our dataset (\ref{SubSec:data_prep}) has a train-validation split of 90\%-10\%. We generate a set of angles and translations that are different from the original dataset to formulate the test set.
 
We perform training using an ADAM \cite{kingma2014adam} optimizer, with decay parameters of ($\beta_{1} = 0.9 $ and $\beta_{2}= 0.999$). The experiments were run with a learning rate of $0.009$ and with $10\%$ decay every 250 epochs. The models were trained with softmax multi-class cross-entropy loss, \textbf{$\mathcal{L}$}, represented in equation \ref{eqn:cross-entropy}.
The training was done on 2 Nvidia RTX 2080 Ti GPUs having 11GB of GPU memory each and 64 GB of RAM.

\begin{equation} \label{eqn:cross-entropy}
    \mathcal{L} = \frac{1}{m}\sum_{m=1}^{M}\sum_{i=0, j=0}^{n_k, n_g}-y_{m, i, j}\log(E_{m, i, j})
\end{equation}

where,
    $y_{m, i, j}$ is the target one-hot encoded label representing emotion class $i\{=0, 1,..n_k\}$ and view-group $j\{=0, 1, ...n_g\}$ for the data sample $m\{=0,1, .., M\}$. $E_{m, i, j}$ is the predicted $softmax$ output probability for data sample $m$ being emotion class $i$ and view-group class $j$.

\subsection{Comparing {\algoname} with other Emotion Classifiers}
We evaluate the performance of our {\algoname} network, against two other emotion classification algorithms \cite{bhattacharya2020step}~\cite{Ewalk}. 
Since the other emotion classification algorithms don't consider the arbitrary view scenario, we compare our results with just single-view data, i.e., skeletal gaits that are directly approaching the RGB-D camera. Table \ref{tab:comp_sv_data} presents these results. The accuracy metrics reported are generated by modifying the equations in \ref{sec:eval_metrics}, for a single view-group (i.e., $n_g = 1$).
\begin{table}[h!]
\centering
\small

    \begin{tabular}{|c|c|}
    \hline
        \textbf{Method} & \textbf{Accuracy (\%)} \\ [0.5ex] \hline
         Venture et al. ~\cite{venture2014recognizing} & 30.83 \\ [0.5ex] \hline
         Daoudi et al ~\cite{daoudi2017emotion} & 42.5 \\ [0.5ex] \hline
         Li et al. ~\cite{li2016identifying} & 53.7\\ [0.5ex] \hline
         Baseline (Vanilla LSTM) ~\cite{Ewalk}  & 55.47 \\ [0.5ex] \hline
         Crenn et al ~\cite{crenn2016body} & 66.2 \\ [0.5ex] \hline
         STEP~\cite{bhattacharya2020step} & 78.24 \\ [0.5ex] \hline
         \textbf{{\algoname}~(ours) }& \textbf{82.4} \\ [0.5ex] \hline
    \end{tabular}
    

    \caption{\textbf{Comparison of {\algoname} with other \sota emotion classification algorithms}: We compare the accuracy (\%) of our {\algoname} network with existing emotion classification algorithms on single-view (facing the camera) data samples. We observe that our network outperforms the current \sota~algorithm by \textbf{~4\%}. Furthermore, our network outperforms the \sota~algorithm across each emotion class. The accuracy numbers reported for ~\cite{Ewalk}, ~\cite{bhattacharya2020step} and {\algoname} are evaluated on the same dataset discussed in section \ref{SubSec:data_prep}. The other methods are evaluated on different datasets.
    }
    \label{tab:comp_sv_data}
\end{table}
\vspace{-0.3cm}
\subsection{Comparing {\algoname} with Action Recognition Models}
As mentioned in section \ref{action_comparison}, action recognition models and emotion recognition models that have inputs as gaits are closely related tasks. Thus, we can evaluate {\algoname} on pre-existing action recognition models by fine-tuning them on the emotion recognition task.
We compare our model with two existing {\sota} action recognition models, (i) Spatial-Temporal Graph convolution networks (ST-GCN) \cite{yan2018spatial}, and (ii) VS-CNN \cite{ji2019large}. These architectures were trained using the datasets \cite{Ewalk, bhattacharya2020step} (discussed in Section \ref{SubSec:data_prep}).

\subsubsection{ST-GCN}\label{stgcn}
The spatial-temporal graph convolution networks \cite{yan2018spatial} perform skeletal action recognition using undirected spatial-temporal graphs for hierarchical representation of skeleton gait sequences. In the original implementation, the spatial-temporal graphs are used in a graph convolution network to detect the action performed through the sequence.

We fine-tune ST-GCN to predict human emotion instead of the actions. The human emotions modeled as a class label for the implementation. 




\subsubsection{VS-CNN}\label{vs_cnn}
One of the major drawbacks of ST-GCN is that it is not tuned for multi-view/arbitrary-view skeletal gait sequences.
View-guided Skeleton CNN (VS-CNN)\cite{ji2019large} approaches this problem by building a dataset that multiple view-points with respect to the human reference frame.
The multiple views are combined into four groups, each consisting of the one-quarter (90 degrees) of the view-points sequences. The action recognition is performed in three stages: (i) a \emph{view-group predictor network} that predicts the view-group $C$ (of 4 view-groups) of the sequence. (ii) a \emph{view-group feature network} that consists of \emph{four} individual networks, based on SK-CNN \cite{skcnn}, for each view-group, and finally, (iii) a \emph{channel classifier network} that combines (i) and (ii) to predict the action label for the skeletal gait sequence.

The VS-CNN also steers away from graph convolutions with an aim to increase the run-time performance of the network. 2D convolutions were observed to be much faster and efficient as opposed to graph convolutions. Hence, the spatial-temporal skeletal gait sequences are transformed into images. 
In our experiment, we tweak the final output of VS-CNN architecture using equation \ref{softmax_calc} to predict human emotions as opposed to actions. The network was trained with a $softmax$ cross-entropy loss function, represented in equation \ref{eqn:cross-entropy}.

The table \ref{tab:comp_sv_data} and figure \ref{fig:model_acc_results} present a comparison of our model against VS-CNN and ST-GCN. We can observe that {\algoname} outperforms the {\sota} action recognition algorithms in both single-view and arbitrary-view skeletal gait sequences.
Also, observe that in table \ref{t_computation}, {\algoname} takes up the least number of model parameters. This is because we perform group convolutions and eliminate Fully Connected layers in our network.
Figure \ref{fig:cm} is a confusion matrix of the predicted vs actual emotion classes of {\algoname}. We can infer from this matrix that our model performs fairly well across all emotion classes with a high accuracy.
Since, the evaluation metrics for \textbf{\textit{socially acceptable}} is not well-defined, we don't report any evaluation on our \textit{emotion-guided navigation planning}.

\begin{table} [h!]
\small
\centering
\begin{tabular}{|c | c | c | c |}
\hline
\multirow{2}{*}{\textbf{View-Groups}} & \multicolumn{3}{c|}{\textbf{Model Parameters}} \\ [0.5ex]
\cline{2-4}
 & ST-GCN \cite{yan2018spatial} & VS-CNN  \cite{ji2019large} & \algoname \textbf{(ours)}\\ [0.5ex] 
\hline
4 &1.4M & 63M & \textbf{0.33M} \\ [0.5ex]
\hline
6 &1.4M & 65M & \textbf{0.5M} \\ [0.5ex]
\hline
8 &1.4M & 68M & \textbf{0.69M} \\ [0.5ex]
\hline
\end{tabular}\\

\caption{\textbf{Comparison of model parameters}: Our \algoname~model has significantly fewer parameters compared to ST-GCN \cite{yan2018spatial} and  VS-CNN~\cite{ji2019large}. This is due to the fact that we use \textit{Group Convolutions (GC)} and eliminate \textit{Fully Connected (FC)} layers in our network.
\vspace{-0.4cm}}
\label{t_computation}
\end{table}

\section{Conclusion, Limitations and Future Work}
We present {\algoname}, a novel group convolution-based deep learning network that takes 3D skeletal gaits of a human and predicts the perceived emotional states \textit{\{happy, sad, angry, neutral\}} for \textit{emotionally-guided} robot navigation. Our model specifically takes into consideration arbitrary orientations of pedestrians and is trained using augmented data comprising of multiple view-groups. We also present a new approach for socially-aware navigation that takes into consideration the predicted emotion and view-group of the pedestrian in the robot's field of view. In doing this, we also define a new metric for computing comfort space, that incorporates constants derived from emotion and view-group predictions. The limitation of our model during inference time is that it is reliant on real-time 3D skeletal tracking. 

In the future, we plan to look at multi-modal cues for emotion recognition. We intend to dynamically compute proxemic constraints using continual feedback in a reward-based training scheme.  We also plan to add higher-level information, with regards to the environmental or cultural context that are known to influence human emotions, which can further improve our classification results. 

\section*{Acknowledgements}
This research was supported in part by ARO Grants W911NF1910069, W911NF1910315, NIST and Intel.

{\small
\bibliographystyle{IEEEtran}
\bibliography{refs}
}
\end{document}